\documentclass[a4paper,12pt,reqno]{article}
\usepackage[utf8]{inputenc}
\usepackage{graphicx,amssymb,amsmath,amsthm,mathrsfs,bbm}
\usepackage{accents}
\usepackage{xcolor}
\usepackage{color, colortbl}
\usepackage{wrapfig}
\usepackage{leftindex}
\usepackage{natbib}
\usepackage[hidelinks]{hyperref}
\usepackage[T1]{fontenc}
\usepackage{textcomp, gensymb}
\usepackage{authblk}
\usepackage{multirow}
\usepackage{caption}
\usepackage{subcaption}
\usepackage{mwe}
\usepackage{geometry}
\usepackage{array}
\usepackage{todonotes}
\usepackage[font={small,it}, labelfont=bf]{caption}
\usepackage{booktabs}
\usepackage{siunitx}

\xdefinecolor{dgreen}{rgb}{0,0.5,0}
\xdefinecolor{dgreenish}{rgb}{0,0.5,0.2}
\xdefinecolor{dred}{rgb}{0.65,0,0}
\xdefinecolor{dpurple}{rgb}{0.6,0,0.6}
\definecolor{Gray}{gray}{0.9}
\definecolor{LightCyan}{rgb}{0.88,1,1}

\providecommand{\keywords}[1]
{
  \small	
  \textbf{\textit{Keywords---}} #1
}

\title{Block Graph Neural Networks\\[1mm] for tumor heterogeneity prediction}
\author[1]{Marianne Ab\'{e}mgnigni Njifon \thanks{Supported by Deutsche Forschungsgemeinschaft GRK 2088.}}
\author[2]{Tobias Weber}
\author[3]{Viktor Bezborodov}
\author[3]{Tyll Krueger}
\author[1]{Dominic Schuhmacher.}
\affil[1]{Institute for Mathematical Stochastics\\ University of G\"{o}ttingen}
\affil[2]{T\"{u}bingen AI Center\\ University of T\"{u}bingen}
\affil[3]{Wroc{\l}aw University of Science and Technology}
\date{}

\begin{document}

\maketitle

\begin{abstract}
Accurate tumor classification is essential for selecting effective treatments, but current methods have limitations. Standard tumor grading, which categorizes tumors based on cell differentiation, is not recommended as a stand-alone procedure, as some well-differentiated tumors can be malignant. Tumor heterogeneity assessment via single-cell sequencing offers profound insights but can be costly and may still require significant manual intervention. Many existing statistical machine learning methods for tumor data still require complex pre-processing of MRI and histopathological data.

In this paper, we propose to build on a mathematical model that simulates tumor evolution (\cite{Ozanski_2017}) and generate artificial datasets for tumor classification. Tumor heterogeneity is estimated using normalized entropy, with a threshold to classify tumors as having high or low heterogeneity. Our contributions are threefold: (1) the cut and graph generation processes from the artificial data, (2) the design of tumor features, and (3) the construction of Block Graph Neural Networks (BGNN), a Graph Neural Network-based approach to predict tumor heterogeneity. The experimental results reveal that the combination of the proposed features and models yields excellent results on artificially generated data ($89.67\%$ accuracy on the test data). In particular, in alignment with the emerging trends in AI-assisted grading and spatial transcriptomics,
our results suggest that enriching traditional grading methods with birth (e.g., Ki-67 proliferation index) and death markers can improve heterogeneity prediction and enhance tumor classification.
\end{abstract}


\section{Introduction}

Tumors arise from an accumulation of genetic changes that ultimately lead to uncontrolled cell growth, which is not halted by the body's natural processes. Tumors can be broadly classified into two categories: malignant and benign.  Malignant tumors are those that have the ability to invade adjacent tissues or to metastasize (extend to nearby or other parts of the human body). These are the tumors that are commonly referred to as cancerous. While benign tumors are typically localized, grow slowly, and do not spread to other parts of the body, they can still cause problems by exerting pressure on surrounding tissues. It is observed that many benign tumors remain benign indefinitely, and most malignant tumors do not originate from benign ones. Some specific types of benign tumors however (adenomas, dysplastic nevi, and actinic keratosis) can be precursors to cancerous ones through a process called "malignant transformation", which is influenced by genetic and environmental factors. Early detection and treatment come therefore as an undeniable necessity. 

Significant efforts have been made to classify tumors (\cite{Berman_2004,Ebata_2021,Marzouka_2018,JGO427}). Common indicators used to characterize the "aggressiveness" of tumors are tumor grades and tumor heterogeneity 
(\cite{Jogi_2012, Dagogo_2017, Schmidt_Efferth_2016}). Tumor grade is a measure of the visual appearance of cells under a microscope, as originally defined by the National Cancer Institute of the United States of America in 2022. First, a sample of tissue or specimen is extracted by biopsy. Then, the tumor grade is determined in a laboratory by observing how differentiated the cells look. Typically, the grade is assessed based on the degree of cellular differentiation, mitotic activity, and nuclear atypia. Cell differentiation is an important factor in determining prognosis and treatment options, and its impact can vary depending on the specific tumor type. For example, Higher-grade tumors (Grade III or IV) are generally more aggressive and have a worse prognosis compared to lower-grade tumors (Grade I or II). However, standard grading alone is not sufficient to determine malignancy. \cite{Giacomelli_Sacco_Papa_Carr_2023} points for example to the fact that some tissues could still be cancerous independently of cell differentiation. This means that some cells can be relatively well-differentiated while still being cancerous. It is specifically the case for actinic keratoses (\citeyear{righi2021metabolomic}). Thus, analyzing tumor heterogeneity, that is specific morphological and phenotypic cell profiles (\cite{ijms17122142}), has become a major area of research and serves as an indicator in predicting its aggressiveness.

We distinguish between inter-tumor heterogeneity and intra-tumor heterogeneity (\cite{PINTO201356}). "Inter-tumor heterogeneity refers to the differences in genetic, molecular, and phenotypic traits of tumors across different individuals or among tumors in separate regions of the same organ. This implies that tumors from the same tissue type can exhibit significant differences. Inter-tumor heterogeneity then introduces potential differences in treatment response among patients with the same type of cancer, which motivates the call for more targeted treatment (\cite{Liu_Dang_Wang_2018}). Intra-tumor heterogeneity refers to the variability of genetic, molecular, and cellular characteristics within a single tumor. It is often driven by clonal evolution, where different "subclones" within a tumor acquire distinct mutations over time. This means that different groups of cells or even individual cells within the same tumor can exhibit significant differences, such as distinct mutations, gene expression patterns, and behavior, which can lead to treatment resistance and relapse within the same patient (\cite{Liu_Dang_Wang_2018}). 
We also distinguish between genetic and non-genetic sources of heterogeneity, a discussion that we leave for the appendix (see Page~\pageref{sec:appendix}).
Cells within the same tumor that share specific genomic markers or modifications can be referred to as subpopulations (\cite{Fisher_2013}) or subclones (\cite{Liu_Dang_Wang_2018}).
This work focuses on intra-tumor heterogeneity which in the rest of the paper we simply call heterogeneity.

Biologists typically resort to single-cell sequencing to reveal the heterogeneity of tumor cells (\cite{Lin_Shen_2022, Wu_2021, Schmidt_Efferth_2016}). It does not directly measure the mutation rate of individual cells. However, it provides information about the overall amount of cells simultaneously undergoing mutations, which can provide a relevant indication of the aggressiveness of a tumor. Despite the thorough understanding of relevant changes at the cell level it brings, single-cell sequencing comes with many limitations. Some of these include premature termination of reverse transcription, high cost, low flux, lack of automation, and sometimes a large number of sample cells are required as starting materials (\cite{Qu_2023}). Furthermore, single-cell sequencing is highly sensitive to technical noise, such as amplification bias and dropout events, which can complicate data interpretation. Additionally, analyzing single-cell data requires sophisticated computational tools to handle the high dimensionality and sparsity of the data, which can be a significant challenge. Various methods have therefore been investigated to mitigate the existing drawbacks.

Statistical methods for example have been proposed to infer the heterogeneity of tumor cells by using sequencing data (\cite{Abecassis_2021, Oesper_2013}). Machine Learning (ML) and statistical methods have also been introduced to predict tumor grades using MRI Scans and histopathology images (\cite{Surov_2018, cnn_mri_book_2018, Park_2023, Prabhudesai_2021, Pulvirenti_2021, Miloushev_2014, Deng_Zhu_2023, Wetstein_2022, Nakamoto_2019,BEREBYKAHANE2020401}). Due to their expressive power, Graph Neural Networks (GNNs) have been increasingly used for tumor classification tasks (\cite{MENG2023107201, Ding_Zhou_2022, Ramirez_2020, AYAZ2023105286, 10.1093/bioinformatics/btad643, Ravinder_2023}).  One of the advantages of ML models is their ability to integrate and analyse complex datasets from various sources and modalities (\cite{kather2019deep, schmauch2020deep, cheerla2019deep, Cai_Poulos_2022, Li_Nabavi_2024, Chen_Lu_2022, Waqas_2024}). One active area of research is the use of the proliferative index to enhance the automation of ML-based tumor grading (\cite{joseph2019proliferation, feng2020automated, kather2019deep, yucel2024automated}).

All the previously mentioned methods require the collection and use of annotated biological data, which can be time-consuming and costly to obtain. Whole slide images, which are used in some of these methods, come with several challenges, including tissue slide-dependent issues (such as scratches on slides, artifacts on pixels, irregularly shaped or fragmented tissues), device-dependent issues (such as characteristics of the optical system used to produce the optical image of the tissue slide), and storage challenges (\cite{basak_wsi}). Moreover, these datasets require extensive pre-processing including sometimes the usage of pre-trained deep learning models for feature extraction resulting in performance that is unfortunately not easily explainable, in the sense that it poses a level of challenge to understand which features played a substantial role in the obtained performance and why.

In this work, we propose a GNN-based approach to model the heterogeneity of simulated tumor data, utilizing handcrafted features. This is achieved in 3 major steps: (1) First, we leverage the tumor simulation model proposed in \cite{Ozanski_2017}, generate tumor cuts, and build graphs to constitute our dataset. (2) Next, we design node features based on the spatial distribution of cells, which are used to characterize our graph nodes. This framework is translation-invariant since it only depends on the relative positions of the cells, not their absolute positions. Some designed features include the use of birth and death markers (3) Then we propose a Graph neural network-based model that we call Block Graph Neural Networks (BGNN) to solve the heterogeneity prediction task. 

The next section of the paper (Section \ref{sec:Dataset_descr}) provides a concise description of how the artificial tumor is simulated. This is followed by a section that details how the generated data is adequately prepared to be passed on to the classification model for its analysis. Next, Section \ref{sec:methodology} presents the BGNN methodology, which is the classification model introduced in this research. The experimental results in Section \ref{sec:results} provide insights into the importance of each designed feature, the performance of the BGNN model, and possible extensions to improve the obtained results. The final section summarizes our findings and suggests future directions for research.

\section{Tumor simulation}\label{sec:Dataset_descr}

\subsection{Presentation of the tumor simulation model}
Recent studies report on the advantages of using artificial data for tumor analysis (\cite{hu2022synthetictumorsmakeai, D’Amico_2023, Cai_Apolinário_2024})
The artificial data considered in this paper is generated from a tumor model proposed in \cite{Ozanski_2017}. Comparisons with similar models are beyond the scope of this paper and thoroughly discussed in \cite{Ozanski_2020}. This model is a spatial birth-and-death process that mimics the dynamics of particles interacting in a three-dimensional system. The rarity of mutation events in DNA coupled with the multiplicity of cells in a human body makes it difficult to directly observe the initiating mutations that would later lead to cancer cells. For these reasons, it is assumed here that the tumor has gone through those critical mutations already, which means that its growth can no longer be limited by the body (either by repair or destruction of the damaged cells). All the nutrients that are necessary for the vital functions of cells are captured through a \emph{local density parameter}. This parameter plays a fundamental role in the total population size: low density favors longevity and proliferation of cells. For each cell in the system, we can either observe a death (removal of the parent cell) or a division (simultaneous removal of the parent cell and creation of two child cells). A cell’s division rate is determined by two parameters: birth efficiency and birth resistance. The probability that an initiated division succeeds depends on success efficiency and success resistance. The time, until a cell dies, is governed by lifespan efficiency and lifespan resistance. These parameters, which apply to isolated particles, are referred to as the intrinsic parameters of the simulation and are used to compute the probability of each event occurring for an individual cell. For each pair of parameters, the efficiency sets the base value of the property for an isolated particle, while the resistance determines beyond which the base value is reduced depending on the local particle density (see Appendix for more details).
Besides those, there also exist parameters that pertain to the entire simulation, named ``global'' parameters. Among those, we mention the \emph{mutation rate}, which plays a major role in the observed tumor heterogeneity. Each newly born cell inherits its parent cell's parameters, with some probability of variation of these original parameters, called mutation probability. Further descriptions of those parameters are provided in the Appendix.
Sets of cells that share the same type of mutation are called subclones or clones. The entire tumor evolution is described by the interplay of the above-mentioned parameters. 

\subsection{Tumor generation}
Running simulations with different sets of parameters results in tumors with distinct growth scenarios. In practice, the tumor growth simulation terminates when either the pre-defined time has elapsed or a pre-specified number of birth events have occurred.
We can extract thin layers from the simulated tumors by analyzing the generated cell history, which we refer to as \emph{tumor cuts}. These tumor cuts can be further partitioned into \emph{tumor patches}. 
From a histopathological perspective, tumor patches are similar to additional cuts that could be performed to ensure a thorough analysis of specific areas, if a more detailed examination is needed. This is also similar to the sectioning of the specimen into smaller areas during the grossing stage. The grossing stage in laboratories involves slicing a specimen into small enough sections and placing them into a cassette for further analysis. This process typically occurs before thin-section extractions.
In our simulation, these tumor patches are smaller overlapping chunks of tumor cuts, which enable us to zoom into specific areas of tumor cuts and virtually increase the data size available for training. Our goal is to classify individual tumor patches into low and high heterogeneity ones,  which will provide insights into the heterogeneity of the tumor. Future works will propose ensemble methods to infer the heterogeneity of the global tumor by analyzing the heterogeneity of the various patches. In what follows, we describe the tumor cut process. 

\subsection{Cut procedure}  
The efficient implementation of the model described above is given by \cite{Ozanski_2020}. The tumor simulation stores each cell event together with the spatial coordinates of each cell, which serve as a unique cell identifier. Let $s = \left(x,y,z\right) \in \mathbb{R}^3$ represent the spatial coordinates of a given cell.
To perform a cut, we select a time window $\Delta t$  during which we mark down the birth and death events and a spatial window $\Delta z$ that specifies the thickness of the cut. 
We use a time window instead of a single time moment because in a real tumor, cell division takes a certain time, whereas in the simulation \cite{Ozanski_2020}, it happens instantaneously. Time here represents the in-simulation time (\cite{Ozanski_2020}), and the thickness is chosen to be 6 spatial units, which is roughly 3-4\% of the total tumor width. This thickness choice appears realistic when considering existing pathology reports. Values of tumor sizes for colorectal cancer cases have been published in \cite{Kornprat_2011} to be in the range of 0.6 cm to 15 cm, and \cite{JGO427} recorded a cut size of 3 to 4 mm. This amounts to a thickness range of 2\% to 66.67\% of the overall tumor width, with a median of 6.67\% (\cite{Kornprat_2011}).

The tumor cut is obtained by considering the set of points $\chi_{\Bar{z}, \Bar{t}}=\{\left(x, y, z\right) | z \in \left[\Bar{z}, \Bar{z}+\Delta z \right], t \in \left[ \Bar{t}, \Bar{t} + \Delta t   \right]\}$. $\Bar{z}$ represents a reference coordinate value on the axis perpendicularly to which the cut will be performed (the $z$ axis is chosen here without any loss of generality) and $\Bar{t}$ represents a reference time. Denote by $\chi$ the set of all cuts $\chi_{\Bar{z}, \Delta t}$ obtained by respectively performing cuts for the reference points $\Bar{z} \in \{0, -6, 6\}$ and $\Bar{t} \in \{40, 60\}$. The spatial locations are also in-simulation coordinates. The chosen thickness (3-4\%) is achieved by setting $\Delta z$ to $3$. We choose $\Delta t=1$, which corresponds to a realistic range of cell division as observed in histopathological images of marker Ki67 (\cite{joseph2019proliferation}).

\begin{figure}[ht!]
    \includegraphics[width=\textwidth]{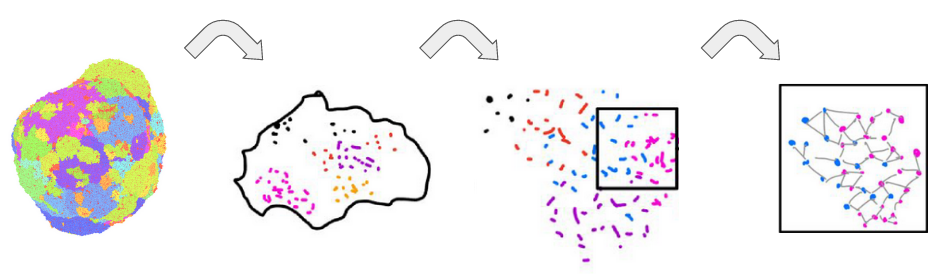}
    \caption{Data generation process. From left to right: (1) A tumor evolution is simulated given a set of global and intrinsic parameters (\cite{Ozanski_2020}); (2) Thin layers called tumor cuts are later prepared from the synthetic tumor; (3) Tumor patches are generated from tumor cuts by randomly selecting positions and collecting all the cells within a designated neighborhood; (4) Graph data are constructed from selected patches, while the remaining patches are discarded.}
    \label{fig:data-generation_proc}
\end{figure}

Figure \ref{fig:data-generation_proc} illustrates the various steps in the artificial tumor and tumor cuts generation process.

\section{Data Preparation}
\subsection{Patch and graph generation/selection.} 
The patch generation process starts by uniformly sampling $100$ positions at random from each $\chi_{\Bar{s}, \Bar{t}} \in \chi$.
Note that there are $6$ cuts in $\chi$.
Then, starting from those center positions, we consider for each $\chi_{\Bar{s}, \Bar{t}} \in \chi$ all the points within a radius of 10 spatial units, which form the tumor patches. After constructing 10-nearest-neighbor graphs from each patch of points, we select all graphs that contain more than 100 edges and more than 100 points for our dataset. For each patch, we calculate its normalised entropy (cf. Section \ref{sec:normalised_entropy}) and discard those graphs whose entropy falls within a small margin of the heterogeneity threshold used for defining the binary classification task (see Section \ref{sec:normalised_entropy}).

\begin{figure}[ht!]
    \includegraphics[width=\textwidth]{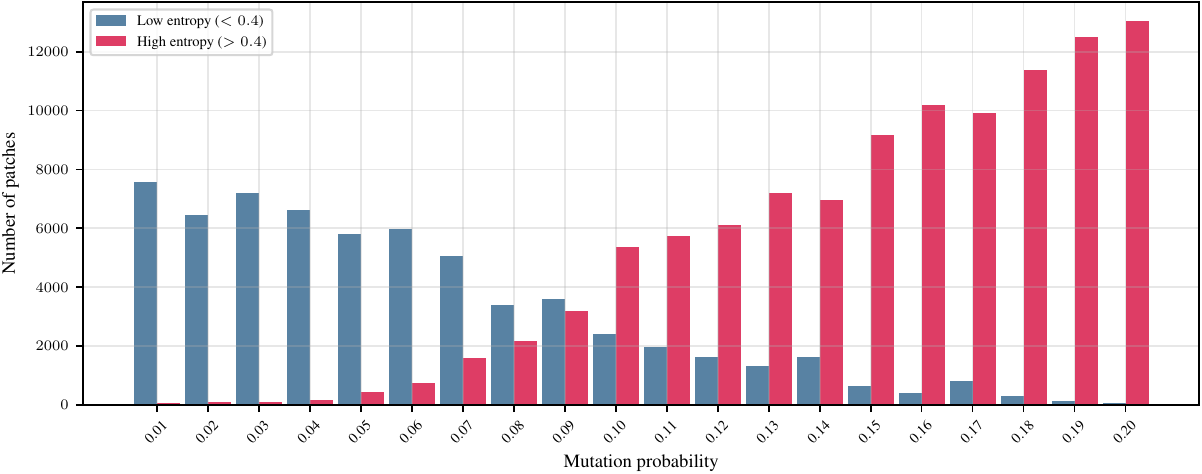}
    \caption{Class label distribution for patches as a function of mutation probability. Class "$0$" corresponds to "low" heterogeneity patches, while class "1" corresponds to "high" heterogeneity patches. We observe that the heterogeneity level of patches increases as the mutation probability increases. On the one hand, when distinguishing between benign and malignant tumors, the increase in heterogeneity level indicates a higher probability of cancer. On the other hand, this heterogeneity can equally help distinguish between different degrees of aggressivity in malignant tumors.}
    \label{fig:data-mutation_prob}
\end{figure}

 Figure \ref{fig:data-mutation_prob} depicts the distribution of patch classes as a function of the mutation probability of the cells. As the mutation probability increases, the likelihood of different subclones appearing also increases, resulting in a higher probability of obtaining heterogeneous patches (red).

\subsection{Node and edge features}
In this section, we describe the design of node features used for tumor heterogeneity modeling. In the following, we consider a single graph $G:=(V, E)$ where $V=\{v_1, \ldots, v_n\}$ denotes the set of all $n$ nodes and $E \subset V \times V$ represents the corresponding set of edges.

\textbf{Local intensity $\lambda(v)$}.
Let $v \in V$ be a node with coordinates $s$, where $s$ represents the spatial coordinates of the node. 
We denote by $\mathcal{N}_v$ the neighborhood of $v$, which is the set of all nodes $u \in V$ that are connected to $v$: $\mathcal{N}_v = \{u \in V | \{v, u\}\in E\} $. An estimator of the local intensity $\lambda(v) $ is obtained by weighting the distances between $v$ and other nodes, and taking the average. More precisely,
\begin{equation}
    \lambda(v) = \frac{1}{\vert \mathcal{N}_v \vert} \sum_{v_j \in \mathcal{N}_v} \exp \left[ -\frac{1}{2} \left(\frac{\|v-v_j\|}{\sigma}\right)^2 \right].
\end{equation}
The weighting reduces the contribution of distant nodes and increases the contribution of close nodes. Intuitively, the local intensity estimates on average the relative influence of neighboring nodes based on their distances to the node of interest.

\textbf{Local density $\rho(v)$.}
The local density (\cite{Ozanski_2020}), similar to the local intensity, aims to estimate the relative influence of nodes with two major differences:
\begin{enumerate}
    \item the local density considers the distances between the node of interest and all the nodes of the tumor,
    \item the kernel used for weighing purposes is a generalized version of the truncated exponential function.
\end{enumerate}
The local density is formally defined as:
\begin{equation}
    \rho(v) =  \sum_{v_j \in V} \kappa_{\rho} \left(\|v-v_j\|\right)
\end{equation}
with 
\begin{equation}
    \kappa_{\rho}\left(w\right) = \begin{cases}
        b \exp{\left[-\frac{1}{\gamma}\left(\frac{w}{\sigma_{\rho}}\right)^{\gamma}\right]} \quad \text{if}\hspace{0.2cm} w<r_{\rho}, \\ 0  \quad \text{otherwise},
    \end{cases}
\end{equation}
where $b$, $\sigma_{\rho}$ and $\gamma$ are respectively scale, width and shape parameters; $V$ %
is the set of all the nodes of the tumor.
The local density is automatically generated during the tumor simulation. The kernel parameters as stated in \cite{Ozanski_2020} are the following: $\gamma=2$, $b=\sigma_{\rho}=1$ and $r_{\rho}\approx 1.517 \sigma_{\rho}$.

It is worth noting that both the local intensity and local density expressions use kernels that are not normalised, meaning they do not integrate to 1. They are simply used for weighting purposes.

\textbf{Birth binary encoding $\delta_{birth}(v)$.} 
For each node $v \in V$, we define a birth binary encoding feature $\delta_{birth}(v)$, which is assigned a value of 1 if there was a birth event for the cell into consideration, and 0 otherwise (in other words, a cell is marked with the value 1 if it gives birth to at least one other cell).
 The birth binary encoding provides on-site information concerning the birth event of the node of interest during the cut window.

 \textbf{Death binary encoding $\delta_{death}(v)$.} 
We similarly define the death binary encoding to the birth binary encoding, to provide on-site information about the death activity of the node.

\textbf{Cell volume $a(v)$.} 
We consider the Voronoi tessellation of the set of coordinates of the nodes (\cite{Voronoi1908NouvellesAD}), which partitions the subset of $\mathbb{R}^3$ containing the tumor into regions (Voronoi cells) such that all locations closer to a specific node than to any other nodes are contained in the same region.
We denote the volume of the Voronoi cell containing the node $v$ as $a(v)$.  Like the local intensity, the cell volume helps assess the spatial configuration of the graph.

\textbf{Local birth $\lambda_{birth}(v)$.}
Denote by $V_{birth}$, the set of nodes that exhibit a birth event during the cut window, where $V_{birth} \subset V$. 
Similarly to the local intensity, we can estimate the importance of the birth activity of nodes adjacent to a node of interest using the expression:
\begin{equation}
    \lambda_{birth}(v) = \frac{1}{\vert V_{birth} \vert} \sum_{v_j \in V_{birth}} \exp{\left[ - \frac{1}{2} \left( \frac{\|v-v_j\|}{\sigma} \right)^2 \right]}
\end{equation}
A high birth act
ivity does not necessarily correlate with high mutation rates and therefore heterogeneity. Nevertheless, different birth signals around nodes can provide insights into local activities happening at those nodes.

\textbf{Local death $\lambda_{death}(v)$.}
We define the local death intensity as:
\begin{equation}
\lambda_{death}(v) = \frac{1}{\vert V_{death} \vert} \sum_{v_j \in V_{death}} \exp \left[-\frac{1}{2} \left( \frac{\|v-v_j\|}{\sigma} \right)^2 \right]
\end{equation}
where $V_{death}$ is to the set of all nodes $v_j \in V$ that have a death event in the cut window. 

All the above-described features are stored together as one feature vector $h \in \mathbb{R}^{7}$ for the node $v$, which is defined as:

\begin{equation}
    h = \begin{pmatrix}
        \lambda(v)\\ \rho(v)\\ \delta_{birth}(v)\\ \delta_{death}(v)\\ a(v)\\ \lambda_{birth}(v)\\\lambda_{death}(v)
    \end{pmatrix}
\end{equation}
The input matrix $H \in \mathbb{R}^{N\times7}$ is the concatenation of features from all nodes and will be fed as input to the graph neural network described in Section \ref{sec:methodology}. Note that by design, the graph is translation invariant which matches the task at hand.

\textbf{Edge feature}:

In this paper, the only edge features we consider are the Euclidean distances between nodes, that is: 
\begin{equation}
e_{i,j} = \|v_i-v_j\| \hspace{0.2 cm} \forall i=1,\ldots, n \hspace{0.2 cm} \text{and}\hspace{0.2 cm} j=1,\ldots, n \hspace{0.2 cm} \text{with} \hspace{0.2 cm} i\neq j.
\end{equation}

\subsection{Heterogeneity metric: normalised entropy.}
\label{sec:normalised_entropy}

Various metrics have been proposed in the literature for tumor heterogeneity analysis: histogram-based features, Bayesian approaches, intensity-based, and many others (\cite{ O’Sullivan_2005,Just_2014,Xu03042015,Roberts_Newell_2017,Ni02012019, Eloyan_Yue_2020}).
As in \cite{Ozanski_2020}, our target is the normalised entropy metric denoted by $U$, an entropy-based statistic able to assess the heterogeneity of tumor patches. The classification task here is the distinction between high and low entropic tumor patches

First, we define the notion of proportional cluster size. During the data generation process, each subclone of cells can be uniquely identified by a mutation ID. The proportional cluster size $p_i$ of a subclone $i$ is the relative size of the subclone with respect to the total number of cells in the tumor. That is,
 \begin{equation}
     p_i = \frac{n_i}{\sum_i n_i},
 \end{equation}
where $n_i$ is the number of cells with mutation $i$ and $\sum_i n_i$ is the total number of cells in the tumor. 

Now let $N_c$ be the total number of clones in the tumor. The normalised entropy reads:
     \begin{equation}
         U = -\frac{1}{\log_2 N_c}\sum_{i}p_{i}\log_{2}p_{i}.
     \end{equation}
 This metric varies between 0 (when all the cells are part of a single subclone) and 1 (which means that each subclone contains exactly one cell). It expresses how uniform the clone distribution is within the tumor.

We do not discuss the clinical relevance of (normalised) entropy as a heterogeneity measure, and subsequently as an aggressiveness indicator in this paper. However, several authors have investigated this question and revealed the relationship between (normalised) entropy and tumor metabolism, malignancy, patient's response to treatment as well as limitations (\cite{Park_Lim_Nam_Kim_2016,Cheng_2016, Dercle_2017, Henderson_2017, cancers14215235, Costa_2023, Syga_2024}).
Distinction criteria between highly aggressive and less aggressive tumor samples vary greatly, depending on the tumor type, associated data, and the task being solved. \cite{Costa_2023} compares histograms of entropy maps for patients with and without pathological tumor response to chemotherapy, and does not explicitly state any cut-off value. \cite{Failmezger_2022} analysed spatial heterogeneity for bio-markers scoring in patient prognosis prediction. They studied late-stage colorectal cancer (33 patients were in stage IV, and 1 in stage II) and used an entropy threshold of 0.61. It appears from the literature, that lower threshold values should be preferred for lower-grade tumors, while higher cut-offs are indicated for higher-grade tumors. In this work, we choose a normalised cut-off value of 0.4, which also allow for a clear distinction between the 2 heterogeneity classes. However, we highlight the fact that this threshold value is linked to the dataset and not to the methodology. A small class margin of $0.05$ is further applied as a patch selection criteria to enhance the class distinction (we delete patches whose entropy values lie between $0.375$ and $0.425$).

\section{Classification Methodology: the BGNN architecture}\label{sec:methodology}

\begin{figure}[ht!]
    \centering 
    \includegraphics[scale=0.38]{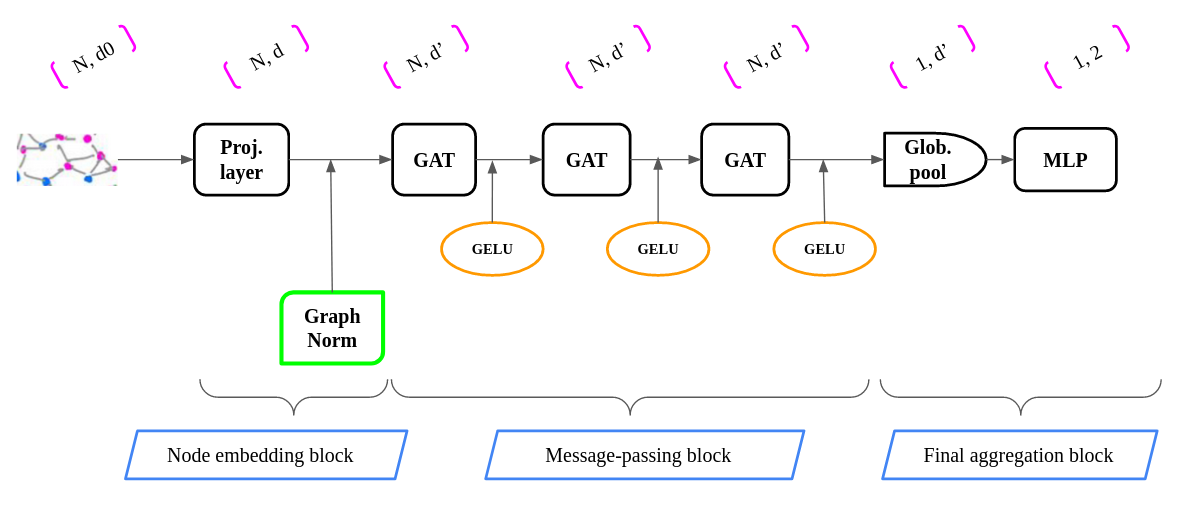}
    \caption[]{BGNN model. Above each layer, the shape of the layer's output is given inside the pink brackets, when considering an input graph of shape $\left(N, d_0\right)$. The 3 blocks of the BGNN are outlined in orange.}
    \label{fig:baseline}
\end{figure}

In what follows, we describe the GNN architecture proposed for the prediction of tumor heterogeneity and we decompose it into three blocks: the node embedding block, the message-passing block, and the final aggregation block. Figure~\ref{fig:baseline} gives an overview of the full architecture.
 
 \vspace{0.5cm}

\vspace{0.5cm}

 \subsection{Node embedding block}
 
 This block serves as an initial encoding of the node features. It is made of a dense projection layer and a graph normalisation layer implemented by "$\texttt{GraphNorm}$" (\cite{Cai_2021}). $\texttt{GraphNorm}$ is a node-wise normalisation operation for graphs that reparametrizes node embeddings by adjusting the proportion of mean feature retained during normalization. Given a node embedding $h_i \in \mathbb{R}^d$ of node $v_i$, for its $j$-th feature, with $j\in\{1, \dots, d\}$, the $\texttt{GraphNorm}$ operation reads:
 \begin{equation}
     \texttt{GraphNorm}\left(h_{i}^{j}\right) = \gamma_j\frac{h_{i}^{j}-\alpha_j\mu_j}{\Hat{\sigma}_j} + \beta_j
 \end{equation}
where $\mu_j = \frac{\sum_{i=1}^n h_{i}^{j}}{n}$ is the feature mean, $\Hat{\sigma}_j =  \frac{\sum_{i=1}^n \left(h_{i}^{j} - \alpha_j\mu_j\right)^2}{n} $ is the feature-adjusted standard deviation. $\gamma_j$ and $\beta_j$ are the learnable parameters of the layer that are also found in other normalization methods, and $\alpha_j$ is a $\texttt{GraphNorm}$ feature-specific learnable parameter that determines the proportion of the mean to subtract for each feature.
 \vspace{0.5cm}
 
 \subsection{Message-passing block}
 
 The goal of this part of the network is to appropriately propagate "relevant" information along the constructed edges. It consists of three nonlinear Graph attention (GAT) layers; which are three GAT layers (\cite{velikovi2017graph}), each of which is followed by a nonlinear activation function $\sigma$. GAT architectures have been greatly solicited in a wide range of applications including graphs. A single nonlinear GAT layer can simply be described as follows: Consider a graph with $n$ nodes, each described by a feature vector $h_i$ with $h_i \in \mathbb{R}^d$ where $d$ is the number of features and $i=1,\ldots, n$.
 The node update of the feature $h_i$ reads:
 \begin{equation}
     h_i^{'} = \sigma \left(\sum_{j \in \mathcal{N}_i} \alpha_{ij} W h_j \right)
 \end{equation}
 $\mathcal{N}_i$ represents 
 the set of indices of all the nodes that belong to some neighborhood of the node $v_i$ (including $i$). $W\in \mathbb{R}^{d'\times d}$ is the layer weight matrix whose parameters are learned during training; $\sigma$ represent the activation function (nonlinear function) and $\alpha_{ij}\in \mathbb{R}$ are the attention coefficients that specify the contribution of node $v_j$'s features to node $v_i$. We use the attention mechanism from \cite{velikovi2017graph}, namely
\begin{equation}\label{attention_coeff_a_ij}
    \alpha_{ij} = \frac{\exp \left(LeakyReLU\left(a^T \left[Wh_i \| Wh_j \|We_{ij}\right] \right)\right)}{\sum_{q \in \mathcal{N}_i} \exp \left(LeakyReLU \left( a^T \left[Wh_i \| Wh_q\|We_{iq}\right] \right)\right)}.
\end{equation}
where $.^T$ indicates the transposed operation,  $\|$ the concatenation, $a\in \mathbb{R}^{3d'}$ is the attention weight vector that is learned during training and $e_{ij}$ denote the feature of the edge connecting the nodes $v_i$ and $v_j$. The activation function
$LeakyReLU\left(t\right) = \max\left(\beta t, t\right), \beta\in \left] 0, 1 \right], t\in \mathbb{R}$ (\cite{maas2013rectifier})
 allows to account for all the node values of the neighborhood during the weight importance calculation, including negative values. It is also possible to endow edges with features and to incorporate the latter in the computation of attention coefficients. 

 In the above-described algorithm, each node $v_i$ gets "attended" to only once (that is, only one importance coefficient $\alpha_{ij}$ is computed for every $j \in \mathcal{N}_i$ ). For this reason, this mechanism is also called single-head attention. We note that, since $v_i$ also belongs to $\mathcal{N}_i$, the relative importance of $v_i$ to itself is computed as well. This is referred to as self-attention. \cite{velikovi2017graph} reported that using multi-head attention could stabilize the learning process of self-attention.
 In the multi-head attention mechanism, several attention mechanisms are performed in parallel on the same node, and the resultant node update is an aggregation of the individual attention heads.

The nonlinear activation function $\sigma$ used here is the Gaussian Error Linear Unit (GELU). GELU's mathematical formulation is given by: 
\begin{equation}
    \operatorname{GELU}\left(x\right) = x.\frac{1}{2}\left[1+erf\left(x/\sqrt{2}\right)\right] \approx  0.5x\left(1+\tanh{\left[\sqrt{2/\pi}\left(x+0.044715x^3\right)\right]}\right)
\end{equation}
where $erf\left(x\right) = \frac{2}{\sqrt{\pi}}\int_{0}^{x}e^{-t^2}dt.$

The effectiveness of GELU was evaluated on computer vision, natural language processing and speech tasks (\cite{hendrycks2023gaussianerrorlinearunits}) and solicited in various modern architectures (\cite{devlin-etal-2019-bert, brown2020languagemodelsfewshotlearners,dosovitskiy2021imageworth16x16words}). An extensive analysis, studying stationarity, differentiability, boundness, and smoothness, was performed by \cite{lee2023geluactivationfunctiondeep}. Here we investigate its performance on GNNs for a classification task involving spatially distributed data.

\subsection{Final aggregation block}
 
The last block of the framework is made of a global average pooling layer across nodes followed by a 2-layer fully connected layer of sizes $2d^{'}$ and $2$, respectively, where $d^{'}$ is the dimensionality of the message-passing block output. 

\subsection{Training:} \label{sec:training}
\begin{figure}[ht!]
        \centering
        \includegraphics[width=\textwidth]{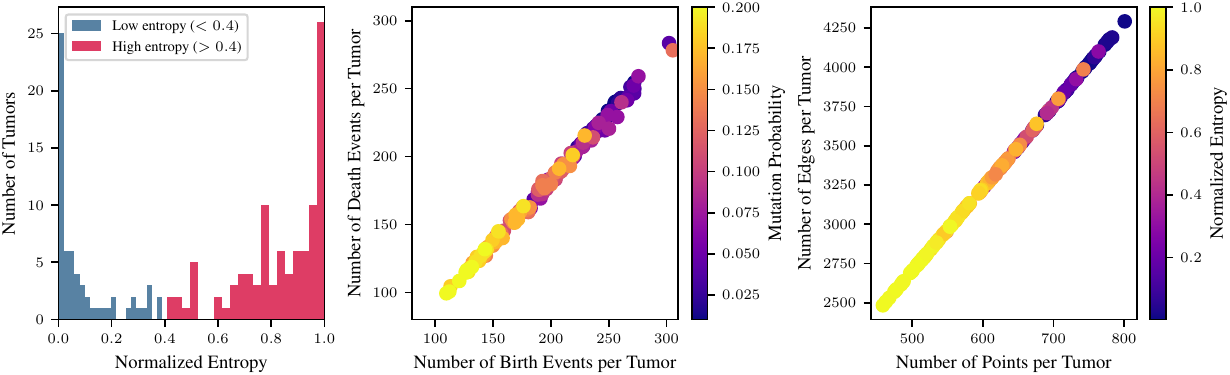}
        \caption[]{Some statistics related to the simulated training data: (a) The histogram shows a prevalence of high entropic patches after applying the selection criteria. A balanced dataset is obtained as described in section~\ref{sec:training}. (b) The birth and death events are not functions of the mutation probability per se. However, higher mutation rates may create more variability among individuals, which is an important aspect of evolutionary dynamics. (c) An increased normalised entropy indicates an increased mutation probability and therefore fewer deaths and fewer birth events. As a result, fewer points are available to generate graphs, which leads to a reduced number of points and edges.} 
        \label{fig:stat_plots_data}
    \end{figure}

We generate a total of $200$ tumors for training ($160$ tumors), validation ($20$ tumors), and testing ($20$ tumors). The parameters used to generate the tumors are given as follows: we respectively have $0.2$ and $0.5$ for the birth efficiency and resistance, $0.9$ and $0.5$ for the success efficiency and resistance, and finally $0.1$ and $0.5$ for the lifespan efficiency and resistance. Each tumor is allowed to freely grow up to at least one million cells. Further details on the simulated data are given in Table~\ref{tab:data_det} and Figure~\ref{fig:stat_plots_data}. The patches obtained for each set (training, validation, test) are re-balanced to match a $50:50$ ratio of high vs low entropic patches. This is achieved in each set by considering the entropic class with the smallest number of patches and discarding the corresponding exceeding number in the other class.

The set of training patches is partitioned into batches and fed as input into the BGNN model in the form of graph data. For each input graph, the model predicts $2$ numbers, probabilities to belong to each entropic class to be fed into the cross-entropy loss function that we recall next.  Let  $T(g)$ be the true probability distribution for a graph tumor patch $g$ and $P(g)$ be its predicted probability distribution. The cross-entropy loss function $\mathcal{L}\bigl(T(g), P(g)\bigr)$ outputs a real number such that:
\begin{equation}
    \mathcal{L}\bigl(T(g), P(g)\bigr) = -\sum_{i=1}^{2} T_i(g) \log{P_i(g)}
\end{equation}
The BGNN is trained to minimize the cross-entropy loss between those predicted values and the target class of each tumor graph
Finally, the accuracy metric is used to score the performance of the model, during the training, validation, and test phase. The classification score here is simply the percentage of correctly classified patches.

\begin{table}[ht!]
\small
\centering
\sisetup{separate-uncertainty}
\begin{tabular}{l 
                S[table-format=3.2(2.2)] 
                S[table-format=3.2(2.2)] 
                S[table-format=3.2(2.2)] }
\toprule
 & {Training} & {Validation} & {Test} \\
\midrule
Number of tumors & 160 & 20 & 20 \\
Number of cuts per tumor & 6 & 6 & 6 \\
Number of patches per cut & 100 & 100 & 100 \\
Average number of cells & 634.26 \pm 86.51 & 632.32 \pm 94.72 & 632.92 \pm 90.00 \\
Average number of births & 198.35 \pm 43.08 & 196.99 \pm 47.43 & 200.25 \pm 40.97 \\
Average number of deaths & 181.82 \pm 40.47 & 181.11 \pm 43.87 & 184.72 \pm 38.32 \\
Average normalised entropy (target) & 0.55 \pm 0.39 & 0.55 \pm 0.40 & 0.55 \pm 0.42 \\
\bottomrule
\end{tabular}
\caption{Statistics of patch generation. Each tumor undergoes six cuts, and we extract 100 patches per cut. The number of cells, births, and deaths in the artificially generated training, validation, and test datasets are averaged over all patches of all tumors for each set. Standard deviations are reported using $\pm$.}
\label{tab:data_det}
\end{table}

\section{Results}\label{sec:results}
In this section, we demonstrate the performance of the BGNN in predicting the heterogeneity of our tumor data. There are two main goals to this experimental study: 1-) Investigate the relevance of each designed feature to the overall classification performance and 2-) Evaluate adaptations of the architecture that could result in more stable learning. 

The implementation is in Python and relies mainly on PyTorch Geometric (\cite{fey2019fastgraphrepresentationlearning}) and PyTorch (\cite{paszke2019pytorchimperativestylehighperformance}).  The hidden dimensions $d$ and $d'$ are set to 64 and the input slope of $LeakyReLU$ ($\beta$) in the attention mechanism is set to $0.2$. To avoid overfitting, we make use of the dropout technique with a dropout value of $0.15$.
The dropout causes the training accuracy to be lower than the validation and test accuracies (Table \ref{tab:feat_acc}). We equally acknowledge potential statistical variation due to the use of a single dataset. 

\subsection{Comparing node features}
\begin{figure}[ht!]
        \centering
        \includegraphics[width=\textwidth]{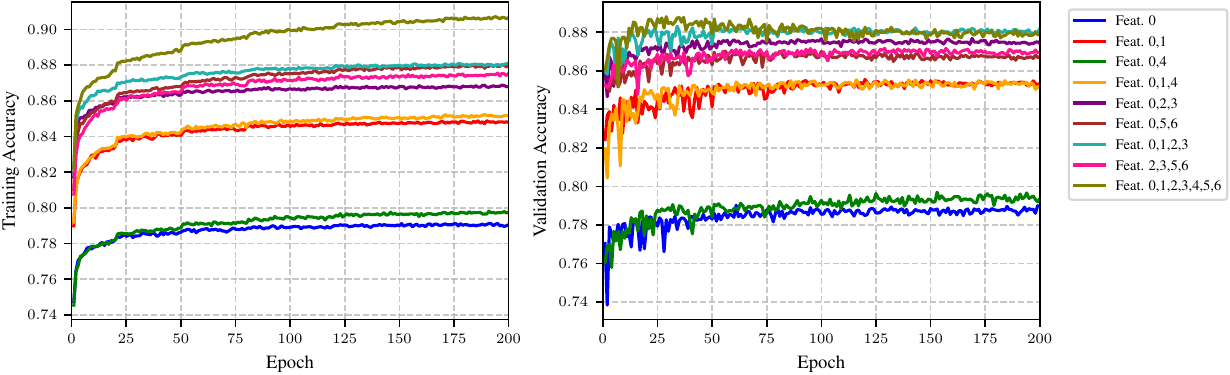}

        \caption[]{Performance of BGNN as a function of node features. The various IDs indicate the node features used during training. ID 0: Local intensity, ID 1: density, ID 2: Local birth, ID 3: Local death, ID 4: Cell volume, ID 5: Birth binary encoding, ID 6: Death binary encoding} 
        \label{fig:mean and std of nets_comp_feat}
    \end{figure}

Here, all models were optimized for $200$ epochs using the ADAM Optimizer (\cite{KingmaB14}), where we adapt the initial learning rate of 1e-3 by a factor of 0.5 every 33 epochs. Figure \ref{fig:mean and std of nets_comp_feat} shows the accuracy of the BGNN model both during training and validation for various feature combinations and Table~\ref{tab:feat_acc} reports the test accuracies. 

First, we observe that training the BGNN model using only the local intensity feature (blue curve, feat:0) yields a test accuracy of $81.05\%$ with a training accuracy of $79.05\%$. This observation indicates that the local spatial structure around the cells adequately informs the BGNN about tumor heterogeneity. Next, we note that augmenting the local intensity feature with the cell volume (green curve, feat:0,4) slightly improves the performance. This indicates that the cell volume does bring additional information to the local intensity feature, probably due to the covering of a somewhat different aspect of the spatial configuration of the graph at the local level. Adding the density feature (yellow curve, feat:0,1,4) improves the accuracy of the predictions. This seems plausible as it provides insight into the spatial structure of the entire tumor graph from a cell's perspective and not only from a local neighborhood around the cell. However, the most impactful features remain those that encode birth-and-death information about cells. In particular, the local birth and death density (purple curve: feat 0,2,3) seems to be more influential than the local birth and death binary encoding (brown curve, feat:0,5,6). Not only do the former achieve a higher test accuracy, but they seem to even generalize better (a lower training accuracy yielding a higher test accuracy), in comparison to the local birth and death intensity. Those features work jointly with the local intensity and the other features to improve the performance of the BGNN (see the higher test accuracy in the presence of the features in the olive line, feat:0,1,2,3,4,5,6).

\begin{table}[ht!]
\small
\centering
\begin{tabular}{l S[table-format=2.2] S[table-format=2.2] S[table-format=2.2]}
\toprule
Features Used & {Train Acc. (\% $\uparrow$)} & {Val Acc. (\% $\uparrow$)} & {Test Acc. (\% $\uparrow$)} \\
\midrule
Feat. 0 & 79.05 & 78.98 & 81.05 \\
Feat. 0,1 & 84.79 & 85.32 & 86.32 \\
Feat. 0,4 & 79.76 & 79.35 & 81.71 \\
Feat. 0,1,4 & 85.17 & 85.24 & 86.73 \\
\rowcolor{LightCyan}
Feat. 0,2,3 & 86.77 & 87.42 & 88.73 \\
\rowcolor{LightCyan}
Feat. 0,5,6 & 87.95 & 86.74 & 88.26 \\
\rowcolor{LightCyan}
Feat. 0,1,2,3 & 88.10 & 88.19 & 89.28 \\
\rowcolor{LightCyan}
Feat. 2,3,5,6 & 87.53 & 86.96 & 88.90 \\
\rowcolor{LightCyan}
Feat. 0,1,2,3,4,5,6 & 90.66 & 87.90 & 89.58 \\
\bottomrule
\end{tabular}
\caption{Classification accuracies of BGNN on the train, validation, and test sets as a function of node features. The various IDs indicate the node features used during training and evaluation: ID 0 (Local intensity), ID 1 (Density), ID 2 (Local birth), ID 3 (Local death), ID 4 (Cell volume), ID 5 (Birth binary encoding), and ID 6 (Death binary encoding). The shaded rows correspond to feature combinations including birth and death information. Higher values indicate better performance.}
\label{tab:feat_acc}
\end{table}

\subsection{Extended architectures: performance on a small-size dataset.}

\begin{figure}[ht!]
    \centering
    \begin{subfigure}[b]{0.185\textwidth}   
        \centering 
        \includegraphics[width=\textwidth]{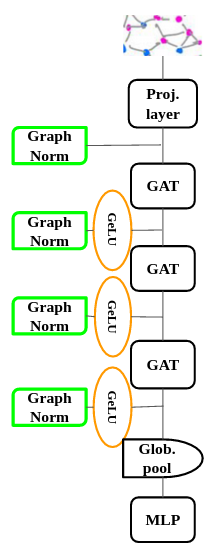}
        \caption[]{baseline + $\texttt{GraphNorm}$ after each non linear GAT layer.}    
        \label{fig:model_2}
    \end{subfigure}
    \hfill
    \begin{subfigure}[b]{0.24\textwidth}   
        \centering 
        \includegraphics[width=\textwidth]{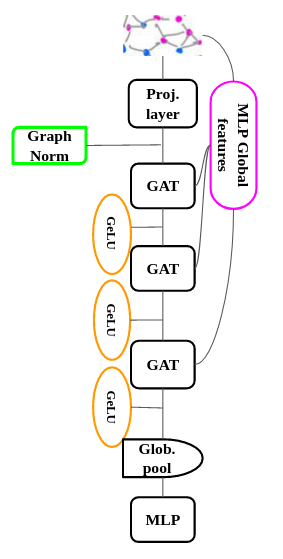}
        \caption[]{ baseline + Input node Global features .}   
        \label{fig:model_3}
    \end{subfigure}
    \hfill
    \begin{subfigure}[b]{0.26\textwidth}   
        \centering 
        \includegraphics[width=\textwidth]{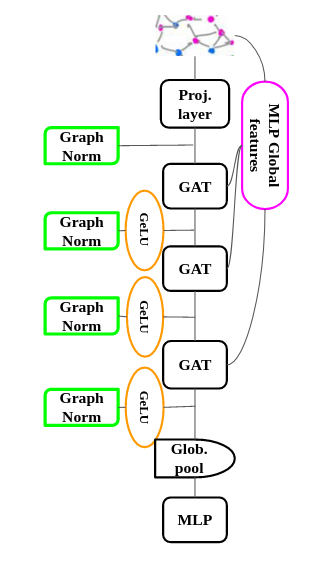}
        \caption[]{baseline + $\texttt{GraphNorm}$ + Input node global features.}
        \label{fig:model_4}
    \end{subfigure}
    \caption[ ]{Investigating different extended architectures.} 
    \label{fig:model_architectures}
\end{figure}

This section investigates the performance of various extended versions of the "vanilla" BGNN model. The experiments here are carried out using all the node features. In total, we consider three different types of extension:
\begin{enumerate}
\item \emph{Feature normalization within the node embedding block (see Figure \ref{fig:model_2}):} Despite the fast convergence and good generalization ability of $\texttt{GraphNorm}$, one drawback of excessive normalization is a potential loss of input information. This analysis helps investigate the effect of normalizing after every layer. 

\item \emph{Global graph feature propagation:} As proposed in \cite{brasoveanu2023extending}, expressive global graph features can increase both the overall expressivity of message-passing graph neural networks and their performance. \cite{brasoveanu2023extending} mainly investigates specialized global features from chemoinformatics, which are endowed with pre-defined meanings for molecular properties. In our case, it appears somewhat difficult to attach a meaning beforehand to global features of local patches. However, we remained interested in studying its effect on the BGNN architecture. With no prior knowledge of the adequate structure of the global feature to use, we proposed to learn one by backpropagation. The global graph feature is hence implemented as a 3-layer neural network (see Figure \ref{fig:model_3}) followed by an average pooling and is, for this reason, referred to as \textit{``MLP Global features''}. These features are concatenated to the input of every non-linear GAT layer and propagated in a densenet-like mechanism (\cite{huang2018denselyconnectedconvolutionalnetworks}). 
\end{enumerate}

Figure \ref{fig:model_architectures} portrays the various combinations we investigate. 

For this set of experiments, we work with a smaller size dataset, where using the same range of simulation parameters, we generate $4614$ tumor patches for training,  $1378$ tumor patches for validation, and $1378$ tumor patches for testing with a $50:50$ ratio of low versus high entropic patches. All models were optimized for 25 epochs using the ADAM Optimizer (\cite{KingmaB14}), where we adapt the initial learning rate of 1e-3 by a factor of 0.4 every 5 epochs. The scores on the test set are summarized in Table~\ref{tab:extended}.

\begin{table}[ht!]
\small
\centering
\begin{tabular}{l S[table-format=2.2] S[table-format=2.2]}
\toprule
Method & {Single-Head Att. (\% $\uparrow$)} & {4-Head Att. (\% $\uparrow$)} \\
\midrule
\rowcolor{LightCyan}
Vanilla BGNN & 88.28 & 89.67 \\
BGNN + GraphNorm & 89.23 & 89.34 \\
BGNN + Global Feat. & 88.09 & 88.23 \\
BGNN + Global Feat. + GraphNorm & 88.79 & 89.45 \\
\bottomrule
\end{tabular}
\caption{Different model accuracies on the small test set. The first column shows the single-head attention results, while the second column shows the 4-head attention results. The vanilla model's performance is highlighted in cyan. Higher values indicate better performance.}
\label{tab:extended}
\end{table}

\begin{figure}[ht!]
    \centering
    \includegraphics[width=\textwidth]{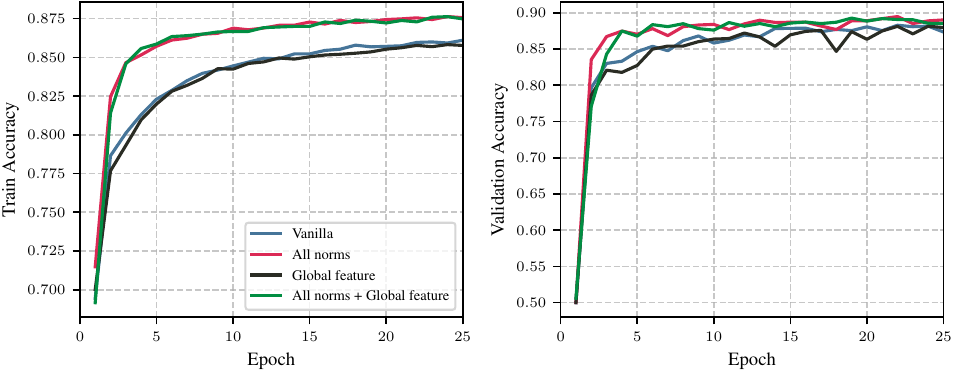}
    \caption[ ]{Training on all 4 tumors with different architecture choices. Vanilla: BGNN baseline, all norm.: BGNN baseline + GraphNorm after each GAT layer, global: BGNN baseline+ Input global feature concatenated at the entry of each GAT layer, glob+norm: BGNN baseline + GraphNorm + Input Global features. All these models use a single-head attention mechanism with a hidden dimension of $64$.} 
    \label{fig:mean and std of nets_mod64}
\end{figure}

The first set of experiments in this section analyses single-head attention performances (see Figure~\ref{fig:mean and std of nets_mod64} and Table~\ref{tab:extended}). 

To begin with, we highlight the fact that, despite the relatively small size of the dataset used, the performance on all the models studied remains above $88\%$ accuracy on the test set. This observation seems to indicate that the BGNN model remains a good model to use irrespective of the size of the dataset, which is a good point in biological applications where some types of illnesses do not present enough specimens for analysis.

Then, comparing the vanilla BGNN (blue curve) and the BGNN+global feature model (black curve), we observe that the MLP Global feature designed here does not help improving the BGNN performance. However, normalizing after every nonlinear GAT layer seems to have added some value to the test set performance as can be seen in the red curve (BGNN+$\texttt{GraphNorm}$) and the green curve (BGNN+global feature+$\texttt{GraphNorm}$).

\begin{figure}[ht!]
    \centering
    \includegraphics[width=\textwidth]{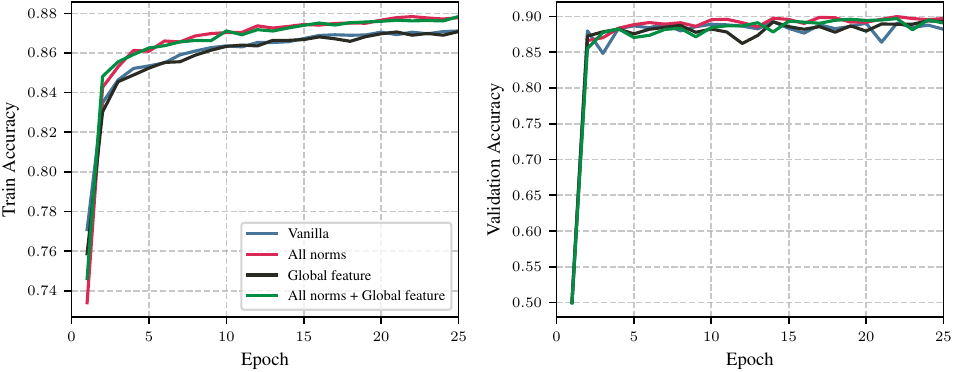}
    \caption[]{Training on all 4 tumors with different architecture choices.  Vanilla: BGNN baseline, all norm.: BGNN baseline + GraphNorm after each GAT layer, global: BGNN baseline+ Input global feature concatenated at the entry of each GAT layer, glob+norm: BGNN baseline + GraphNorm + Input Global features. All these models use a 4-head attention mechanism with a hidden dimension of $64$.} 
    \label{fig:mean and std of nets_mod64_4}
\end{figure}

The last experiment presents the advantage of multi-head attention in this specific task (see Figure~\ref{fig:mean and std of nets_mod64_4} and Table~\ref{tab:extended}). Intuitively, the multi-head attention mechanism makes it possible for each head to focus on different hallmarks during the weight importance computation. We observe that the training procedure is stabilised, and the performance of all 4 models under the 4-head attention is somewhat comparable. The best result on the test set is however achieved on the BGNN baseline while the lowest performing model is the BGNN+global feature.

\section{Conclusion}

In this paper, we present Block Graph Neural Networks (BGNN), a GNN-based approach to predict the heterogeneity of tumors. Analysing the proportion of subclones within the same tumor is useful in at least two classes of problems: (1) to distinguish between benign and malignant tumors, (2) to distinguish between highly aggressive and low aggressive malignant tumors.  Solving those tasks empowers clinicians to administer adequate treatment to patients. Assembling datasets for these classification tasks comes with many costs, and many sophisticated models lack explainability despite their performance. For these reasons, we propose to use a mathematical model of tumor growth to artificially generate various tumor data.
We model the heterogeneity using normalised cross-entropy. The classification is achieved here by proposing a GNN-based framework that leverages the spatial distribution of tumor cells, as well as birth and death information about the said cells.

Our contributions were threefold. Firstly, we described a cut and graph generation algorithm, inspired by biopsy procedures, to produce a training, a validation and a test dataset from the tumor simulation developed in \cite{Ozanski_2020}. Secondly, we proposed a few handcrafted features that can represent available information about each tumor cell (symbolically indexed as a graph node in our formalism). The importance of each proposed feature is studied in the experimental section, where we investigate their impact on the classification performance. Finally, we built the BGNN, a model that consists of 3 main blocks: the node embedding block for feature encoding, the message-passing block for information propagation, and the final aggregation block for the last prediction stage. We also study state-of-the-art extension schemes available in the literature to assess their relevance to the specific heterogeneity classification task.

Experimental results suggest that the local structure around cells is sufficiently represented by either the local intensity feature or the cell volume. More importantly, it appears that among the proposed features, the most influential are those carrying birth and death signals, especially the local birth and death density features. This observation was interesting, given that the proliferative index, which reflects the growth or division activity of cells, is used in different contexts but not  in analysing the heterogeneity in space to conclude about the genetic diversity in traditional procedures. This is an active research area withe emerging single-cell genomic tools. In agreement with recent advances in Automated AI-based tumor classification using proliferation markers, we argue that these hallmarks could boost model performance in distinguishing between malignant and benign tumors, as well as between highly aggressive and low aggressive malignant tumors. By investigating various extensions to the baseline, we confirm that applying the $\texttt{GraphNorm}$ operation more often can sometimes lead to better results, especially on small-size datasets. The vanilla BGNN remains however a good baseline model to work with, especially when the multi-head attention scheme is used. We also do not rule out the possibility of achieving better performance with the global feature strategy, provided that we find an excellent global feature scheme for this purpose.

The positions of cells represent the crucial information that was used from our artificially generated dataset and are typically available in real-world datasets (at least in metadata). In the future, it will therefore be interesting to test both the model and the features on real-world datasets. It may also be worth analyzing the influence of various cut schemes on the classification performance.

\clearpage

\section*{Appendix}\label{sec:appendix}
\subsection*{Intra-tumor heterogeneity taxonomy by source}

We distinguish between genetic and non-genetic sources of intra-tumor heterogeneity (\cite{Shlyakhtina_Moran_Portal_2021}).

Genetic heterogeneity refers to the phenomenon where a single phenotype or disorder is caused by variations in different genes or genetic loci. This can manifest as allelic heterogeneity, where different mutations within the same gene cause similar phenotypes, or locus heterogeneity, where mutations in different genes can produce similar clinical outcomes. Genetic heterogeneity can also arise due to clonal evolution, where mutations accumulate over time, leading to distinct subpopulations within a tumor(\cite{Ng_Turner_Robertson_Flygare_Bigham_Lee_Shaffer_Wong_Bhattacharjee_Eichler_et_al._2009}).

Non-genetic sources of intra-tumor heterogeneity can be subdivided into 5 groups. 
 
There is phenotypic heterogeneity which refers to the variation in observable traits (phenotypes) among individuals who have the same or similar genetic mutation differences. It can arise from can arise from differences in cellular differentiation, microenvironmental factors, and stochastic gene expression. This results in distinct disease severity, age of onset, symptoms, or clinical presentation due to factors such as genetic background, environmental influences, or epigenetic modifications even when individuals share the same genetic basis for a condition (\cite{Grayson_Wang_Aune_2011}).
 
Then comes, epigenetic heterogeneity which is the variation in epigenetic marks (such as DNA methylation, histone modifications, and non-coding RNA expression) between cells, tissues, or individuals, even when they share the same genetic sequence. It promotes potential differences in gene expression patterns, which contribute to diverse cellular behaviors, phenotypes, and responses to environmental stimuli or therapies (\cite{Feinberg_Irizarry_2010}). Noe that epigenetic changes can be reversible and are influenced by factors like hypoxia, immune responses, and inflammation.
 
Next, we have cell behavioral heterogeneity which encapsulates the variability in behavior and functionality observed among individual cells within a homogeneous population (\cite{KLEIN20151187}). Potential sources of cell behavioral heterogeneity include differences in cell signaling, metabolism, and response to environmental cues.
 
Finally, we have Functional and metabolic heterogeneity.  Metabolic heterogeneity indicates the variance in metabolic processes among cells within a population, even when they are genetically identical (\cite{Carthew_2021}),  
while functional heterogeneity describes the dissimilarities in functional capabilities among cells, which can include variations in their ability to perform specific tasks, such as secretion of proteins, proliferation rates, or responses to signals (\cite{Tyurin-Kuzmin_Karagyaur_Kulebyakin_Dyikanov_Chechekhin_Ivanova_Skryabina_Arbatskiy_Sysoeva_Kalinina_et_al._2020})
 
Given our interest in assessing the aggressiveness of tumors from the division of groups of cells into clones and therefore from their phenotypical traits, this paper focuses on modeling phenotypic heterogeneity.

\subsection*{Summary of the artificial tumor generation model}

This section proposes a quick overview of the tumor model with the aim of more precisely presenting the intrinsic parameters introduced in Section~\ref{sec:Dataset_descr}. The full description of the tumor model is given in \cite{Ozanski_2020}. For each cell in the system we observe either a death event or a birth event. A successful birth gives rise to two new daughter cells while a failed birth simply removes the parent cell. The catastrophic death inherent to the failed birth is independent of the previous death event possible for each cell. In what follows, we describe these two processes in detail.

The parameters in the system are described via the local density introduced in Section~\ref{sec:Dataset_descr}. 
The local density $\rho_i$ of a cell $i$ is given by:
\begin{equation}
    \rho_i = \sum_{j \in \Gamma} \kappa_{\rho} \left(\|i-j\|\right),
\end{equation}
where $\Gamma$ is the set of all the cells in the system and $\|i-j\|$ is the Euclidean distance between cells $i$ and $j$. 

The kernel $\kappa_{\rho}$ simply serves as a distance weighting function and is not a probability kernel. $\forall w >0$, it is defined by:
\begin{equation}
    \kappa_{\rho}\left(w\right) = \begin{cases}
        s_{\rho}\exp{\left[-\frac{1}{\gamma_{\rho}}\left(\frac{w}{\sigma_{\rho}}\right)^{\gamma_{\rho}}\right]} \quad \text{if}\hspace{0.2cm} w<r_{\rho}, \\ 0  \quad \text{otherwise}.
    \end{cases}
\end{equation}
where $s_{\rho}, \sigma_{\rho}$ and $\gamma_{\rho}$ are respectively scale, width and shape parameters. The exact values they take in the simulation are $\gamma_{\rho}=2, \sigma_{\rho}=s_{\rho}=1$ to obtain a truncated Gaussian. The cutoff is chosen to be $r_{\rho}=1.517$, such that values smaller than $0.1s_{\rho}$ are truncated to 0.

Based on this local density, the model computes the probability of occurrence of the two possible events (division and death) for each cell. %
The division of a cell involves the removal of the parent cell (original cell) with a concurrent generation of %
two new daughter cells. One of the new cells is %
created at the same location as the parent cell whereas the position of the second child cell is sampled uniformly at random from a standard normal distribution centered at the parent's position. This division process only happens if an additional division test is a success.

Denote by $i$ the dividing cell under consideration. The result of the division test is then a coin-flip with a success probability
\begin{equation}
    s_i\left(\rho_i\right) = s_i f_s\left(\rho_i/\Tilde{s}_i\right)
\end{equation}
where $s_i$ is the success  efficiency, $\Tilde{s}_i$ is the success resistance and $f_s$ is the associated kernel and defined as follows:
\begin{equation}
    f_s\left(w\right) = \exp{\left[-\frac{1}{\gamma_s}\left(\frac{w}{\sigma_s}\right)^{\gamma_s}\right]}
\end{equation}
with $\gamma_s=2$, $w>0$, and $\sigma_s$ is the scaling parameter for success resistance.

If the test succeeds, the parent cell $i$ gives rise to two new cells. We refer to this as a birth event. The birth rate $b_i\left(\rho_i\right)$ is defined:
\begin{equation}
    b_i\left(\rho_i\right) = b_i f_b\left(\rho_i/\Tilde{b}_i\right)
\end{equation}
where $b_i$ is the birth  efficiency, $\Tilde{b}_i$ is the birth resistance and $f_b$ is the associated birth kernel and defined as follows:
\begin{equation}
    f_b\left(w\right) = s_b\exp{\left[-\frac{1}{\gamma_b}\left(\frac{w}{\sigma_b}\right)^{\gamma_b}\right]}
\end{equation}
with $\gamma_b=2$ and just as in the death rate computation, $s_b$ and $\sigma_b$ are the scaling parameters for birth efficiency and birth resistance respectively.

If the test fails, then even if the next event for the parent cell $i$ was originally a birth, this changes immediately to a death. In this case, no additional child cell is created and the parent cell $i$ is removed.

As mentioned above, in addition to a birth event, the other possible event for a parent cell $i$ is a death event. Note that this is different from the catastrophic death event that arises after a failed division. 
The death rate $d_i\left(\rho_i\right)$ of a cell $i$ is inversely proportional to is lifespan rate $l_i\left(\rho_i\right)$ and computed as:
\begin{equation}
    d_i\left(\rho_i\right) = \frac{1}{l_i\left(\rho_i\right)} = \left[s_ll_if_l\left(\frac{\rho_i}{\Tilde{l}_i}\right)\right]^{-1}
\end{equation}
where $f_l\left(x\right)$ is the lifespan kernel function defined as: 
\begin{equation}
    f_l\left(w\right) = \exp{\left[-\frac{1}{\gamma_l}\left(\frac{w}{\sigma_l}\right)^{\gamma_l}\right]}
\end{equation}
In these equations, $l_i$ is the lifespan efficiency, $\Tilde{l}_i$ is the lifespan resistance, $s_l$ and $\sigma_l$ are the scaling parameters for lifespan efficiency and lifespan resistance respectively, and $\gamma_l=2$.

All the efficiency parameters ($b_i, l_i, s_i$) and resistance parameters ($\Tilde{b}_i, \Tilde{l}_i, \Tilde{s}_i$) are defined between $0$ and $1$. 
They are also called intrinsic parameters. The scaling parameters ($s_b, \sigma_b, s_l, \sigma_l, s_i$) take arbitrary values and allow to scale the intrinsic parameters to arbitrarily large ranges. 

After a birth event, the newly generated cells do not automatically inherit the parameters of the parent cells. In the event of a mutation, the intrinsic parameters of the newly generated cells are sampled uniformly at random from the following probability density function:
\begin{equation}
    f\left(x|x_0\right)=\begin{cases} \frac{1}{\min\left(x_0+s,1\right)} \quad \text{if}\quad 0<x<\min\left(x_0+s,1\right) \\
    0 \quad \text{otherwise}
    \end{cases}
\end{equation}
where $x_0$ is the intrinsic parameter of the parent cell, and $s$ represents the parameter that defines the maximum permissible increase.
All remaining parameters of the daughter cells are sampled uniformly at random from a uniform distribution that spanned from 0 to the value of the mother cell increased by 0.1.

\end{document}